\pdfoutput=1

\documentclass[11pt]{article}

\usepackage{EMNLP2022}

\usepackage{times}
\usepackage{latexsym}
\usepackage{times}
\usepackage{latexsym}

\usepackage{inconsolata}

\usepackage[T1]{fontenc}

\usepackage[utf8]{inputenc}

\usepackage{microtype}
 \usepackage{enumitem}
\usepackage{amsmath}
\usepackage{graphicx}
\usepackage{amsmath}
\usepackage{amssymb}
\usepackage{booktabs}
 \usepackage{multirow}
 \usepackage{multicol}
\usepackage{times}
\usepackage{epsfig}
\usepackage{graphicx}
\usepackage{amsmath}
\usepackage{amssymb}
\usepackage{bm}
\usepackage{booktabs}
\usepackage{multirow}
\usepackage{threeparttable}
\usepackage{subfigure}
\usepackage{float}
\usepackage{diagbox}

\newcommand{\tabincell}[2]{\begin{tabular}{@{}#1@{}}#2\end{tabular}} 

\title{LogicSolver: Towards Interpretable Math Word Problem Solving with Logical Prompt-enhanced Learning}

\author{{Zhicheng Yang$^{1,2}$\thanks{~~Zhicheng Yang and Jinghui Qin are contributed equally to this work.}, Jinghui Qin$^{3\ast}$}, Jiaqi Chen$^{2,4}$, Liang Lin$^2$ \and Xiaodan Liang$^{1,2}$\thanks{~~Xiaodan Liang is the corresponding author.} \\
{$^1$Shenzhen Campus of Sun Yat-sen University, $^2$Sun Yat-sen University,}\\
{$^3$Guangdong University of Technology,$^4$Dark Matter AI Inc.} \\
{\normalsize\texttt{\{yangzhch6,qinjingh\}@mail2.sysu.edu.cn}} \\ 
{\normalsize\texttt{linliang@ieee.org,}} {\normalsize\texttt{\{jadgechen,xdliang328\}@gmail.com}}
}

\begin{document}
\maketitle
\begin{abstract}
Recently, deep learning models have made great progress in MWP solving on answer accuracy. However, they are uninterpretable since they mainly rely on shallow heuristics to achieve high performance without understanding and reasoning the grounded math logic. To address this issue and make a step towards interpretable MWP solving, we first construct a high-quality MWP dataset named InterMWP which consists of 11,495 MWPs and annotates interpretable logical formulas based on algebraic knowledge as the grounded linguistic logic of each solution equation. Different from existing MWP datasets, our InterMWP benchmark asks for a solver to not only output the solution expressions but also predict the corresponding logical formulas. We further propose a novel approach with logical prompt and interpretation generation, called LogicSolver. For each MWP, our LogicSolver first retrieves some highly-correlated algebraic knowledge and then passes them to the backbone model as prompts to improve the semantic representations of MWPs. With these improved semantic representations, our LogicSolver generates corresponding solution expressions and interpretable knowledge formulas in accord with the generated solution expressions, simultaneously. Experimental results show that our LogicSolver has stronger logical formula-based interpretability than baselines while achieving higher answer accuracy with the help of logical prompts, simultaneously. The source code and dataset is available at \href{https://github.com/yangzhch6/InterMWP}{https://github.com/yangzhch6/InterMWP}.

\end{abstract}

\section{Introduction}
\begin{figure}[t] 
	\centerline{\includegraphics[width=0.99\linewidth]{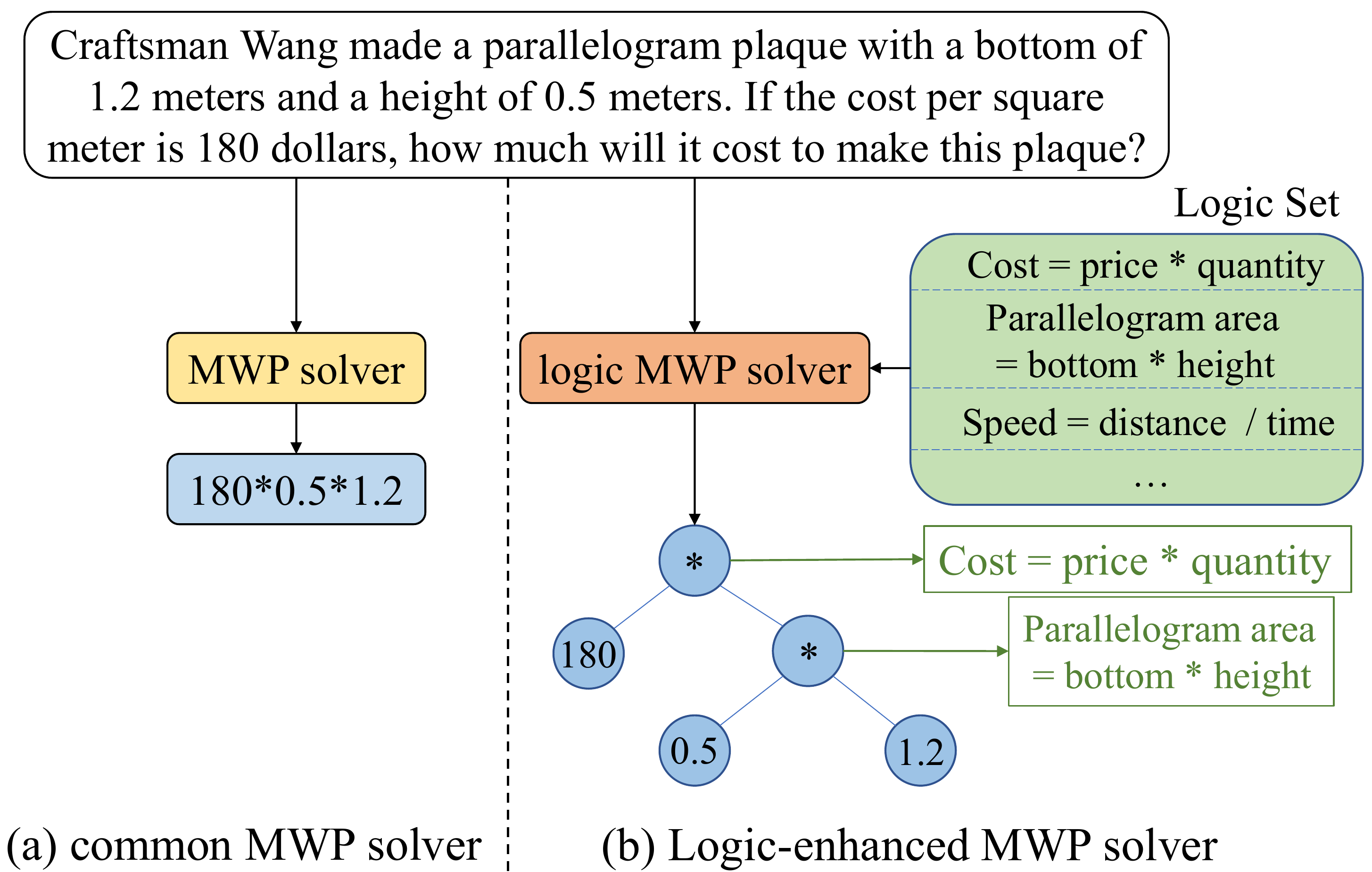}}
	\caption{Common MWP dataset v.s. InterMWP dataset. Compared with the common MWP datasets, InterMWP requires a solver to predict expression tree and the corresponding linguistic logic formulas simultaneously for improving the interpretability of a solver.}
	\vspace{-3mm}
	\label{fig:example}
\end{figure}

Automatically math word problem (MWP) solving is a challenging task in natural language processing since it aims to transform a concise narrative rich in mathematical relationships into a solution equation, as illustrated in Figure \ref{fig:example} (a).
Recently, the task of MWP solving automatically has attracted a lot of research attention.
Several deep-learning-based approaches~\cite{dns,cass,seq2tree,trnn,sau-solver,ns-solver} have been proposed and made great progress for MWP solving. However, as shown in Figure \ref{fig:example}(a), current models treat MWP as a seq2seq task, ignoring the interpretability. The grounded logic in the problem consists of two algebraic knowledge formulas: $cost = quantity \times price$ and $parallelogram\; area = bottom \times height$ where quantity is equal to parallelogram area in this MWP, as shown in Figure \ref{fig:example}(b). Without logical reasoning, it is difficult for an MWP solver to explain why such an equation should be generated as the solution. There are two main reasons for the current dilemma: 1) The lack of relevant and easily exploitable interpretable MWP datasets. 2) Current models mainly rely on shallow heuristics to achieve high performance and lack grounded math logic reasoning, as shown in \citet{patel-etal-2021-nlp}.

To overcome this dilemma and make a step towards interpretable MWP solving, we propose a novel high-quality interpretable MWP dataset called InterMWP consisting of 11,495 annotated samples and 210 different logic formulas based on algebraic knowledge. In our InterMWP dataset, each solution equation is annotated with interpretable logic formulas in a tree structure as the grounded logic of each solution equation. As shown in Figure \ref{fig:example}(b), each inner node is annotated with an interpretable algebraic knowledge formula which represents the grounded logic for the subtree with the current node as the root ancestor. With these logic annotations, our InterMWP asks for a solver to not only output the solution equation but also output the logic formulas simultaneously when the current predicted node is an inner-node (operator) during expression reasoning. Therefore, an MWP solver developed on InterMWP can output a solution equation while generating a reasonable formula-based interpretation. We use answer accuracy from prior works \cite{dns,seq2tree,graph2tree}, together with formula accuracy and logic accuracy we proposed in Section \ref{metric} to evaluate the model's solving ability and interpretability.


To leverage mathematical logic knowledge and empower an MWP solver with interpretability, we further present a novel framework named LogicSolver which extracts mathematical logic knowledge as logical prompts to improve the semantic representations of MWPs and enhance the ability of explanation generation. In our LogicSolver, we design a logic formula retriever to first extract logic prompts consisting of logic formulas highly-correlated with current MWP. Then, the logic prompts will be concatenated with the problem text as the input and drive the MWP model to produce the solution equation. Finally, to obtain the logic formulas-based explanation, we propose a logic generator to predict logic formulas for each inner-node of the solution expression tree. Experimental results show that our LogicSolver has stronger logical formula-based interpretability than baselines while achieving higher answer accuracy with the help of logical prompts, simultaneously.

In this work, our contributions can be summarized in the following three folds:
\begin{itemize} [leftmargin=*]
    \vspace{-2mm}
	\item We construct a high-quality interpretable MWP dataset InterMWP for interpretable MWP solving. In our InterMWP, there are 11,495 MWPs and each solution equation is annotated with interpretable logical formulas. 
	\vspace{-2mm}
	\item We propose a powerful framework named LogicSolver to incorporate mathematical logic knowledge through logical prompt-enhanced learning for enhancing problem understanding while empowering models with interpretability. To the best of our knowledge, this is the first work to study prompt-enhanced learning in MWPs.
	\vspace{-2mm}
	\item We achieved 2.1\%, 2.9\%, and 9.5\% improvement on answer accuracy, formula accuracy, and logic accuracy respectively. Experimental results on InterMWP show that our LogicSolver has strong logical formula-based interpretability which achieves higher answer accuracy simultaneously.
\end{itemize}




\section{Related Work}

\subsection{Math Word Problem Solving} 
In recent years, deep learning-based models~\cite{dns,cass,mathdqn,trnn,seq2tree,stackdecoder,tsrmd,graph2tree,sau-solver,ns-solver} have shown impressive performance in solving MWPs by automatically learning to directly translate a problem text into an expression without any hand-crafted feature design. \citet{dns} make the first attempt to apply a vanilla sequence to the sequence (seq2seq) model. \citet{cass} improved their work by introducing a copy and attention mechanism. \citet{seq2tree} propose a tree-structure decoder to decode expressions in prefix order. Furthermore, \citet{graph2tree} improved problem representation by fusing a graph encoder. \citet{hong2021smart} propose a situation model for algebra story problems. \citet{ns-solver} propose auxiliary tasks to improve problem representation and the ability to predict common-sense constants. \citet{nums2t} achieved better performance by incorporating numerical values into a sequence-to-tree network and applying a numerical properties prediction mechanism. \citet{unbiasedmwp} propose an unbiased dataset and a dynamic target selection (DTS) strategy to eliminate the solving bias.
However, all these models lack grounded math logic reasoning and interpretability so they can not give a reasonable explanation corresponding to the generated expression. To overcome these issues, we build a novel high-quality interpretable MWP dataset and propose a linguistic logic-enhanced framework for generating expression trees and their corresponding formula-based interpretations.  

\subsection{Prompt-enhanced Learning}
Prompting PLMs for few-shot learning has become a popular learning paradigm. PET~\cite{PET-a, PET-b} transfer text classification problems to cloze-style problems while using manually defined prompts to provide additional task guidance. \citet{P-few} propose a pipeline for automating prompt generation to facilitate prompt discovery. \citet{P-what} extract prompt from the training corpus. Besides that, \citet{knowprompt} injects latent knowledge contained in relation labels into the prompt for relation extraction. \citet{knowledgeable} also incorporate external knowledge into a verbalizer to improve and stabilize prompt-tuning for text classification. Although \citet{knowprompt, knowledgeable} incorporate knowledge into PLMs, they mainly focus on the shallow representation.
Unlike these works, we train a model to select prompt automatically from a manually designed prompt set which summarizes the mathematical knowledge needed to solve the math word problems.

\subsection{Interpretability of MWP Solvers}
Although the prior statistical models with hand-crafted features can be thought of as interpretable due to the clear alignments between inputs and outputs, recently proposed deep learning approaches present new challenges to model interpretability of MWP solvers~\cite{huang-etal-2016-well}. \citet{liang-etal-2018-meaning} used pattern matching to increase the robustness and interpretability of MWP solvers. \citet{Amini2019MathQATI} propose operation-based formalisms to improve the interpretability. \citet{GSM8K} propose an MWP dataset called GSM8K which annotates the explanation for each step. But they do not summarize the mathematical knowledge in explanation. Besides, \citet{map_declare} also propose declarative rules which govern the translation of natural language to math expressions and presents a framework that learns to select the relevant declarative knowledge for each operation of the expression. 
Different from these works, we propose to predict linguistic math logic involving real-world knowledge along with expression construction so that an MWP solver can explain the grounded reason about the expression generation with linguistic logic formulas.

\section{InterMWP}
\subsection{Data Collection}
Most existing datasets for math word problem solving mainly consist of 4 attributes: problem id, problem text, solution equation, and final answer, such as Math23K~\cite{dns}, MaWPS~\cite{koncel-kedziorski-etal-2016-mawps}, HMWP~\cite{sau-solver}, and CM17K~\cite{ns-solver}. Since there is no annotated explanation for solving equations, an MWP solver is incapable to produce an explanation grounded in the generated equation.
To make a step towards interpretable MWP solving, we construct a high-quality interpretable MWP dataset called InterMWP to empower an MWP solver with the ability of interpretation to reason out solution equations and produce corresponding explanations for the generated equations.   
Excepting from the attributes mentioned above, we add the extra interpretable formula-based tree-structure annotation into the dataset so that we can force an MWP solver to not only output solution equation but also give out grounded logic formulas on the operators, thus endowing the MWP solver with certain interpretability.  

To collect InterMWP, we sampled 8260 examples randomly from Math23K and crawled another 3,235 examples from a web bank\footnote{https://damolx.com/} to increase data diversity. In total, there are 11,495 examples collected in InterMWP. For each example, we first transferred the sequence solution equations to tree equations following the method in \citet{seq2tree}. The annotation procedure can be referred to Appendix \ref{procedure} in our supplemental materials. 
\begin{figure}[t] 
	\centerline{\includegraphics[width=1.0\linewidth]{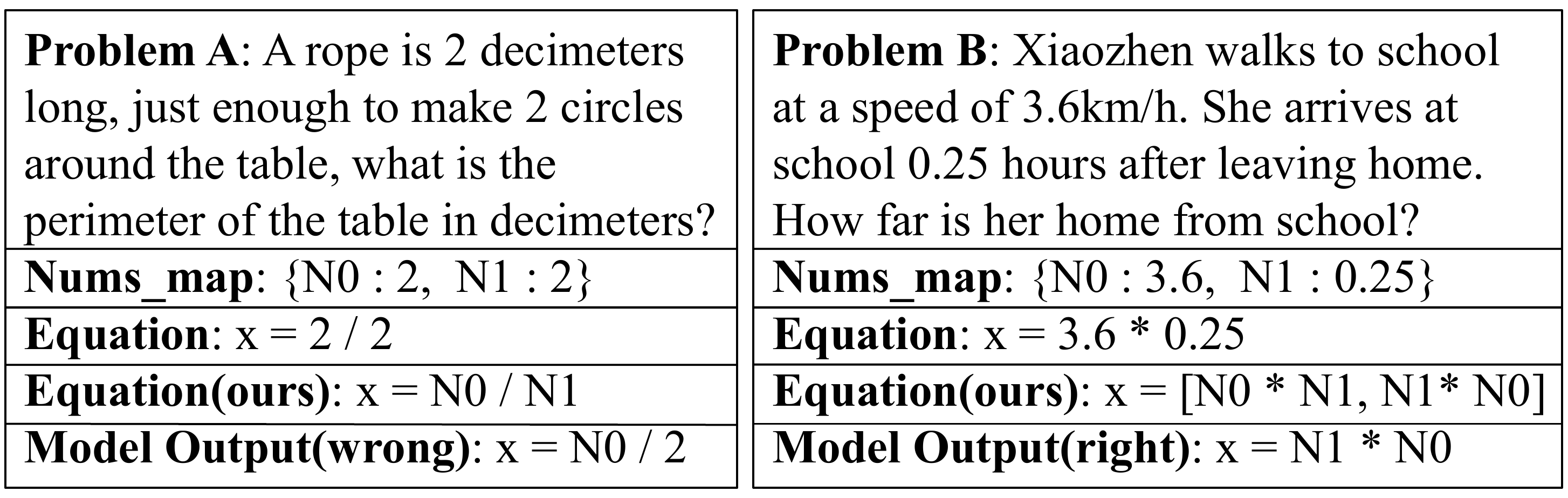}}
	\caption{Some example comparisons between former MWP benchmarks and InterMWP benchmarks.}
	\vspace{-3mm}
	\label{fig:data_drawback}
\end{figure}
\begin{figure*}[t] 
	\centerline{\includegraphics[width=0.85\linewidth]{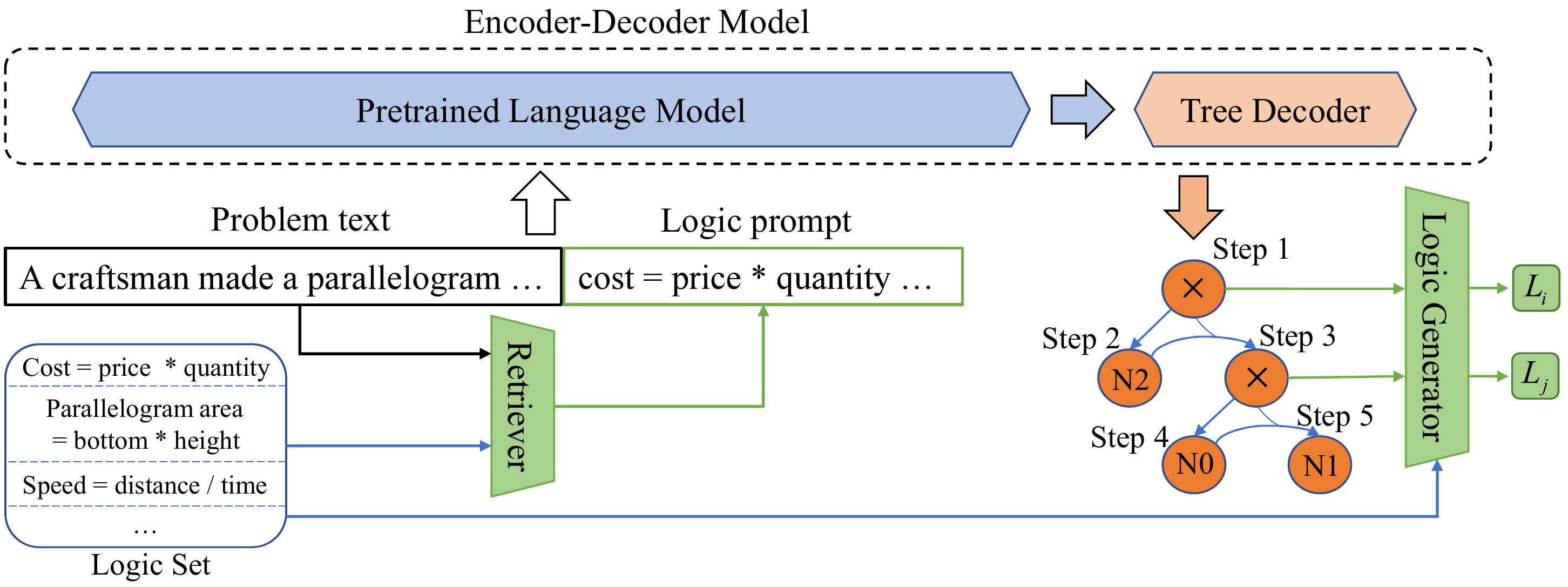}}
	\caption{The design of our proposed LogicSolver. First, we train a logic retriever to extract highly-correlated logical formulas as prompts to solve the MWPs. The retriever takes the problem text and the logic formulas as input and outputs the matching score for each logic formula. Second, we select the top $K$ related logic formulas as prompts and concatenate them with problem text as the input of the encoder while the decoder output solution expression in prefix order. Finally, a logic generator is deployed to select the logic formulas for each operator in the solution expression.} 
	\label{fig:arch}
\end{figure*}
During annotating, we mainly focus on the logic formulas which involve real-world knowledge such as $cost = quantity * price$, $speed = distance / time$, etc. For the basic and simple mathematical logic knowledge described in \citet{map_declare}, we use $common$-$sense\; step$ as the logical formula. The logic formulas involving real-world knowledge are grouped into four main categories: Common-sense, Geometry, Physical, and Finance as shown in Table \ref{tab:lf}. In total, there are 210 formulas summarized in InterMWP. Some examples are illustrated in Table 1 in our supplemental material. It is worth noting that each token in the annotated logic formula can represent the logic semantic of the corresponding node in the tree expression. Taking the logical formula $cost = price * quantity$ of the root node in Figure \ref{fig:example} as an example,  $cost$ is the logical meaning of the root node ("how much will it cost") while $price$ and $quantity$ can represent the logic semantics of its left and right nodes, respectively. 


\subsection{The other superiorities of our InterMWP}
Except for the explainable logic formulas, the other superiorities of our InterMWP can be mainly summarized in the following two points:

\noindent\textbf{a) Formula variables disambiguation}: As the prior MWP datasets such as Math23K~\cite{dns}, Alg514~\cite{alg514} and MAWPS~\cite{koncel-kedziorski-etal-2016-mawps} only provide a numeric expression for each problem, the reference to the variables in the formula may be ambiguous. A data example of such formula ambiguous is problem A shown in Figure \ref{fig:data_drawback}, the original method in \citet{dns} cannot map the two numbers `2' in the equation to different positions in the problem. We overcome this shortcoming by mapping between numbers in the problem and numbers in the solution equation manually during the procedure of annotating logic formulas.

\noindent\textbf{b) The complete solution set for each MWP in the test set}: The former metrics to evaluate the accuracy of an MWP solver mainly rely on the answer accuracy, but an MWP solver may output a right answer by generating a wrong formula. As shown in  Figure \ref{fig:data_drawback}, for problem A, an MWP solver can obtain the correct answer by generating an error constant number `2'. Besides, for problem B,  the generated equation does not match the original equation although they are essentially the same. To overcome this shortcoming, we generate equivalent solution equations as many as possible for each MWP in the test set so that we can measure the ability of an MWP solver better.


\section{LogicSolver}

\subsection{Overview}
As shown in Figure \ref{fig:arch}, there are three main collaborative components in our proposed LogicSolver to solve an MWP and give out the corresponding logic formula-based explanation simultaneously. For each MWP, we first deploy a logic formula retriever to select the top-$K$ highly-correlated logic formulas as logic prompts for prompt-enhanced solving. Then, the logic prompts will be concatenated with the problem text as the input and drive the MWP model to produce a solution equation. Finally, to obtain the logic formulas-based explanation, we deploy a logic generator to predict logic formulas for each inner-node (operator) of the solution expression tree.





\subsection{Logic Formulas Retrieval}
It should be helpful for MWP solving if we can inject the semantics of the logic formulas grounded in MWPs into an MWP solver since the logic formulas grounded in MWPs denote the grounded math relationships. Therefore, to inject logic formulas into an MWP solver to improve the ability of semantic representation and reasoning, we train a retriever to match the logic knowledge in our InterMWP. Our retriever takes a problem text and the 210 logic formulas we summarized as the input and outputs the matching score for each logic formula.

BERT~\cite{bert} is an efficient pre-trained language model for text encoding, so we employ a Chinese BERT pre-trained with whole word masking~\cite{chinese-bert-wwm} as our encoder, denoted as $\mathrm{BERT}_R$, for learning the semantic representations of text. To encode the problem text $P$ and logic formula set $F = [F_1,F_2, \cdots, F_{T}]$ where $T$ is the size of the logic set, we pass them to the retriever $\mathrm{BERT}_R$ and average the feature outputs from the last hidden layer to obtain the corresponding semantic embeddings as follows:
\begin{equation}\label{formula-1}
\begin{aligned}
p & = \mathrm{mean}(\mathrm{BERT}_R(P)) \\
f_i & = \mathrm{mean}(\mathrm{BERT}_R(F_i))\\
\end{aligned}
\end{equation}

Then, a scoring module $\mathrm{Score}_R$ is deployed to rate each logic formula $F_i$ as follows:
\begin{equation}\label{formula-2}
\begin{aligned}
\mathrm{Score}_R(p, f_i) &= \mathrm{v_s^Ttanh}(\mathrm{W_s}[p, f_i]) \\
s_i &= \mathrm{Score}_R(p, f_i) \\
\end{aligned}
\end{equation}
where $\mathrm{v_s}$ and $\mathrm{W_s}$ are trainable parameters.




Given the dataset $\mathcal{D}$, for each data sample $(P, L_f) \in \mathcal{D}$, where $L_f = [l_{f_0}, l_{f_1}, \cdots, l_{f_T}]$ is a 0-1 vector, and $l_{f_i}$ indicate whether the logic formula $F_i$ is used in solving the problem $P$, and $T$ is the size of logic set, we minimize the following loss function for training the retriever:
\begin{equation}\label{formula-3}
\mathcal{L}_r = \mathrm{log}(1+\sum_{l_{f_i} = 1}e^{s_i}) + \mathrm{log}(1+\sum_{l_{f_j} = 0}e^{-s_j})
\end{equation}

For the positive logic formulas, we expect a higher score, and for the negative logic formulas, the opposite is true.

\subsection{Logical Prompt-enhanced MWP Solving}
\label{logical prompt}



To solve an MWP, we follow the encoder-decoder structure. For the encoder, we choose the Chinese BERT pre-trained with whole word masking \cite{chinese-bert-wwm}, denoted as $\mathrm{BERT}_E$, for learning the MWP representation. For decoder, we employ the goal-driven tree-structure decoder, denoted as $\mathrm{GTS}$, following the previous work \cite{seq2tree} to generate solution expression in prefix order, as shown in Figure \ref{fig:arch}.

To conduct logical prompt-enhanced learning with highly-correlated logic formulas, for a problem $P$, we first select the top-$K$ logic formulas based on their matching scores $\{s_0, s_1, \cdots, s_T\}$ as prompts. Then we concatenate the selected $K$ logic formulas with $P$ and pass it into the encoder:
\begin{equation}\label{formula-4}
c, q_{root} = \mathrm{BERT}_E([P, F_{\Omega_K}])
\end{equation}
where $c$ is the encoder's last hidden layer of all tokens, and $q_{root}$ which will be used as the root node's goal vector is the $\mathrm{[CLS]}$ token embedding of the last hidden layer.    

For each node in the expression tree, the goal-driven tree-structure decoder $\mathrm{GTS}$ takes the goal vector $q$ and the token-level embedding $c$ as the input. The decoder $\mathrm{GTS}$ first uses $q$ to predict token $\hat{y}$ of the current node. If the predicted token is a mathematical operator, the goal will be decomposed into two sub-goals which will be passed to the corresponding sub-trees. Otherwise, the goal will be simply realized by the predicted numeric value or constant quantity. The final output of $\mathrm{GTS}$ is the node tokens $\hat{Y}$=$\{\hat{y_1}, \hat{y_2}, \cdots, \hat{y_n}\}$ of the solution expression in prefix order and the corresponding goal vector of each node $Q$=$\{q_1, q_2, \cdots, q_n\}$, $n$ is the solution expression length.
\begin{equation}\label{formula-5}
\hat{Y}, Q = \mathrm{GTS}(c, q_{root})
\end{equation}

Given the dataset $\mathcal{D}$=$\{$$(Y_0,P_0)$, $(Y_1,P_1)$, $\cdots$, $(Y_N,P_N)$$\}$, we minimize the following loss function during training the encoder-decoder model:
\vspace{-4mm}
\begin{equation}\label{formula-6}
\mathcal{L}_d(\hat{Y}_i|P_i) = -\sum_{t=1}^{E_i} \mathrm{log}(\mathrm{prob}(y_t|q_t,P_i))
\end{equation}
where $\hat{Y}_i$ is the predicted expression for $P_i$, $E_i$ is the number of tokens of the solution expression of problem $P_i$, and $\mathrm{prob}(y_t|q_t,P_i)$ is computed by distribution computation function in $\mathrm{GTS}$.

\subsection{Explanation Generation}
\label{logical explanation}
To empower the MWP solver's interpretability, we propose a logic generator to take the operator's hidden embedding $q$, problem text $P$, and the logic set $F$ as the input and predict which linguistic logic formula can explain the decision on the current operator. We deploy a BERT model denoted as $\mathrm{BERT}_L$ to encode each logic formula and problem text.  
\begin{equation}\label{formula-7}
\begin{aligned}
p^L & = \mathrm{BERT}_L(P) \\
f_i^L & = \mathrm{mean}(\mathrm{BERT}_L(F_i)), F_i \in F \\
\end{aligned}
\end{equation}
where $p^L$ and $f_i^L$ denote the problem's token embedding for $P$ and the logic formula embedding for $F_i$ in the logic generator respectively.

To choose a reasonable explanation for the decision on expression generation, we leverage the attention-based scoring mechanism to select an appropriate logic formula as the explanation. Given an operator's embedding $q$, a context vector $c^L$ is obtained by attending $q$ with problem token embedding $p^L$ with the help of the attention mechanism \cite{attention}:
\begin{equation}\label{formula-8}
c^L = \mathrm{Attention}(q, p^L)
\end{equation}
Then, a scoring module denoted as $\mathrm{s}_L$  is deployed to output the unnormalized log probability:
\begin{equation}\label{formula-9}
\mathrm{s}_L(F_i|q,c^L) = \mathrm{v_L^Ttanh}(\mathrm{W_L}[q,c^L,f_i])
\end{equation}
where $\mathrm{v_L}$ and $\mathrm{W_L}$ are trainable parameters.



Finally, the normalized probability $\mathrm{prob}(F_i|q,c^L,F)$ over logic formula set $F$ is computed with the softmax function:
\begin{equation}\label{formula-10}
\mathrm{prob}(F_i|q,c^L,F) = \frac{exp(\mathrm{s}_L(F_i|q,c^L,F))}{\sum_{j}exp(\mathrm{s}_L(F_j|q,c^L,F))}
\end{equation}
Our logic generator selects the logic formula $y^F$ with the highest probability as the explanation for the decision on operator generation: 
\begin{equation}\label{formula-11}
y^F = \mathop{\arg\max}\mathrm{prob}(F_i|q,c^L,F) 
\end{equation}

The training objective of the logic generator is as follows:
\begin{equation}\label{formula-12}
\begin{aligned}
\mathcal{L}_L(\hat{Y}^F_i|P_i) = -\sum_{t=1}^{E^F_i} \mathrm{log}(prob(y^F_i|q_t,c^L_t,P_i))
\end{aligned}
\end{equation}
where $\hat{Y}^F_i$ is the predicted expression for $P_i$, $E^F_i$ is the number of operators of the solution expression of the problem $P_i$.

\section{Experiments}
\subsection{Experimental Setup}
\noindent\textbf{Datasets.} We conduct experiments on our InterMWP and Math23K \cite{dns} in the train-valid-test setting. For Math23K, we train the logic retriever on our InterMWP and then use the pre-trained logic retriever to extract logic prompts for the MWPs in the Math23K. 


\vspace{1mm}
\noindent\textbf{Baselines.} We compare our LogicSolver with the following state-of-the-art models:  \textbf{Math-EN}~\cite{seq2et}: a seq2seq model with equation normalization for reducing target space.  \textbf{GROUPATT}~\cite{group-attn}: a math word problem solver borrowing the idea of multi-head attention from Transformer~\cite{transformer}. \textbf{GTS}~\cite{seq2tree}: a tree-structured neural network in a goal-driven manner to generate expression trees. \textbf{Graph2Tree}~\cite{graph2tree}: an enhanced GTS with quantity graph. \textbf{GTS(BERT)}: a strong baseline we constructed by replacing RNN encoder with BERTEncoder\cite{bert} in GTS.

\vspace{1mm}
\noindent\textbf{Evaluation Metric.} \label{metric}
We use three metrics to measure the problem solving ability and interpretability of the models. 
\begin{itemize}[leftmargin=*]
\vspace{-2mm}
\item Following prior works \cite{dns,seq2tree,graph2tree}, we use \textbf{answer accuracy} as one of the evaluation metrics: if the calculated value of the predicted expression tree equals the true answer, it is thought as correct. 
\vspace{-2mm}
\item However, answer accuracy will overestimate the ability of reasonable expression generation of an MWP solver, so we also introduce \textbf{formula accuracy} to evaluate whether the generated expression is one of a set of reasonable expressions that we annotate an MWP by listing all possible and reasonable solution equations manually in the test set. 
\vspace{-2mm}
\item Moreover, to measure the effectiveness of the output linguistic logic, we introduce \textbf{logic accuracy}: Given the dataset $\mathcal{D}$=$\{$$(Y_0,Y^F_0,P_0)$, $(Y_1,Y^F_1,P_1)$, $\cdots$, $(Y_N,Y^F_N,P_N)$$\}$ where $Y_i$ denotes solution expression, $Y^F_i$ denotes the target linguistic logic formulas, and $P_i$ is the problem text. For an MWP, if the predicted solution expression $\hat{Y}_i$ is correct and the whole predicted linguistic logic $\hat{Y^F_i}$ is equivalent to the target linguistic logic, we consider this logic formula-based explanation is correct. The formula for computing logic accuracy is as following below: 
\end{itemize}
\vspace{-3mm}
\begin{equation}\label{formula-13}
\begin{aligned}
\mathrm{logic\ acc} = \frac{1}{N}\sum_{i=1}^N(\hat{Y}_i = Y_i)(\hat{Y^F_i} = Y^F_i)
\end{aligned}
\end{equation}

\noindent\textbf{Implementation Details.} We use Pytorch\footnote{http://pytorch.org} to implement our model on Linux with an NVIDIA RTX3090 GPU card. We add the [NUM] token to BERT's vocab and convert all numbers in problem text to the [NUM] token.
For the training of the logic generator in LogicSolver, we only select the MWPs which can be fitted in the train set of our InterMWP as the training data. We use goal vectors in GTS-decoder models \cite{seq2tree, graph2tree} as the embedding for solution expression tokens in the logic generator, and select RNN's hidden states in RNN-decoder models \cite{dns, group-attn} as solution expression tokens embedding. More details can be referred to Appendix \ref{details}.

\subsection{Main Result}
\begin{table}[htbp]
\centering
\resizebox{0.99\linewidth}{!}{
\begin{tabular}{l|c|c|c}
\toprule
Model & answer acc & formula acc & logic acc\\ 
\midrule
Math-EN\cite{seq2et} & 63.9 & 60.2 & 43.2 \\
Group-Attn\cite{group-attn} & 64.2 & 60.8 & 44.5  \\
GTS\cite{seq2tree} & 70.5 & 66.1 & 57.2  \\
Graph2Tree\cite{graph2tree} & 71.0 & 66.7 & 57.9 \\ 
NS-Solver\cite{ns-solver} & 71.2 & 66.8 & 57.6 \\
GTS(BERT) & 80.3 & 76.8 & 66.5 \\ 
\textbf{LogicSolver(ours)} & \bf{82.4} & \bf{79.7} & \bf{76.0}\\
\bottomrule
\end{tabular}
}
\caption{The answer acc, formula acc, and logic acc on our InterMWP.}
\vspace{-3mm}
\label{tab:mr}
\end{table}


\noindent \textbf{The Performance on InterMWP.} The results on our InterMWP are shown in Table~\ref{tab:mr}. With the logic prompt-enhanced, the answer accuracy can be improved from 80.3\% [GTS(BERT)] to 82.4\% [LogicSolver(Ours)]. Similarly, the formula accuracy and the logic accuracy also are improved from 76.8\% [GTS(BERT)] to 79.7\% [LogicSolver(Ours)] and 66.5\%[GTS(BERT)] to 76.0\% [LogicSolver(Ours)] respectively. This shows the effectiveness of our proposed prompt-enhanced learning for MWP solving. 
We also evaluate the performance on the samples contain top-10 logic formulas, the results are shown in Table \ref{tab:top_10_acc} of Appendix \ref{exp_logic}.



\begin{table}[htbp]
\small
\centering
\resizebox{0.9\linewidth}{!}{
\begin{tabular}{l|c|c}
\toprule
& GTS(BERT) & GTS(BERT)+Logical Prompts\\ 
\midrule
answer acc & 82.8 &  \textbf{83.4} \\
\bottomrule
\end{tabular}
}
\caption{Experimental results on math23K. The GTS(BERT)+Logical Prompts denotes the GTS(Bert) model enhanced with logical prompts.}
\vspace{-3mm}
\label{tab:math23k}
\end{table}

\noindent\textbf{The Performance on Math23K.} We also conduct the experiment on Math23K. We apply the pre-trained logic retriever on InterMWP to retrieve logic formulas for Math23K, and then conduct logical prompt-enhanced learning for GTS(Bert). The results are shown in Table~\ref{tab:math23k}. The performance of answer accuracy increases from 82.8\% to 83.4\%. This improvement shows the strong generalization of our proposed logical prompt-enhanced learning on other MWP datasets even with the logic prompts based on the InterMWP. 


\subsection{The Performance of the Logic Retriever}
We use Recall, Precision, and F-10 to quantify the performance of the logic retriever on our InterMWP under selecting top 1-4 logic formulas, as shown in Table \ref{tab:retriever}. Corrected logic prompts are very important for improving the MWP solver. To retrieve as many correct logic prompts as possible and decrease the effect of error logic prompts, the recall rate is more important than the precision for logic retrieve. Therefore, we use F-10 score, rather than F-1 score, to measure the balanced performance of the logic retriever.  From Table \ref{tab:retriever}, we can observe that selecting top-3 logic formulas as prompts can achieve a better trade-off between Recall and Precision.


\begin{table}[htbp]
\small
\centering
\resizebox{0.9\linewidth}{!}{
\begin{tabular}{c|c|c|c|c}
\toprule
\diagbox{Indicators}{Selection} & top 1 & top 2 & top 3 & top 4\\ 
\midrule
Recall & 0.642 & 0.948 & 0.985 & 0.991 \\
Precision & 0.557 & 0.411 & 0.284 & 0.215 \\
F-10 & 0.641 & 0.935 & \textbf{0.962} & 0.956 \\
\bottomrule
\end{tabular}
}
\caption{Performance of retriever on InterMWP.}
\vspace{-3mm}
\label{tab:retriever}
\end{table}

\subsection{Logical Prompts Design}

We study the effects of different prompt designs and the effects placement position of logic prompts in the input by conducting three experiments: \vspace{-2mm}
\begin{enumerate}
    \item \textbf{Random Selection}: the logic prompts are choosen randomly.\vspace{-2mm}
    \item \textbf{Retrieve+Ahead}: the logic retriever is deployed for logic prompts retrieving and the logic prompts are put in front of the MWP.\vspace{-2mm}
    \item \textbf{Retrieve+Behind}: the logic retriever is deployed for logic prompts retrieving and the logic prompts are put behind the MWP.\vspace{-2mm}
\end{enumerate}
The results are shown in Table~\ref{tab:r-ab}. From the results, we can know that the best result is obtained under the \textbf{Retrieve+Behind} setting by retrieving top-3 logic prompts.


\begin{table}[htbp]
\centering
\resizebox{0.99\linewidth}{!}{
\begin{tabular}{c|c|c|c|c}
\toprule
\diagbox{Retriever}{Selection} & top 1 & top 2 & top 3 & top 4\\ 
\midrule
Random Selection & 81.1 & 80.4 & 81.5 & 80.6 \\
Retrieve+Ahead & 81.9 & 82.0 & 80.9 & 81.2 \\
Retrieve+Behind & 81.3 & 81.8 & \textbf{82.4} & 81.2 \\
\bottomrule
\end{tabular}
}
\caption{Answer accuracy of different logical prompt designs for LogicSolver (Random-Score denotes the strategy of randomly scoring each logic formula, Ahead and Behind denote the position of prompt).}
\vspace{-3mm}
\label{tab:r-ab}
\end{table}


\begin{figure*}[htb] 
    \vspace{-3mm}
	\centerline{\includegraphics[width=1.0\linewidth]{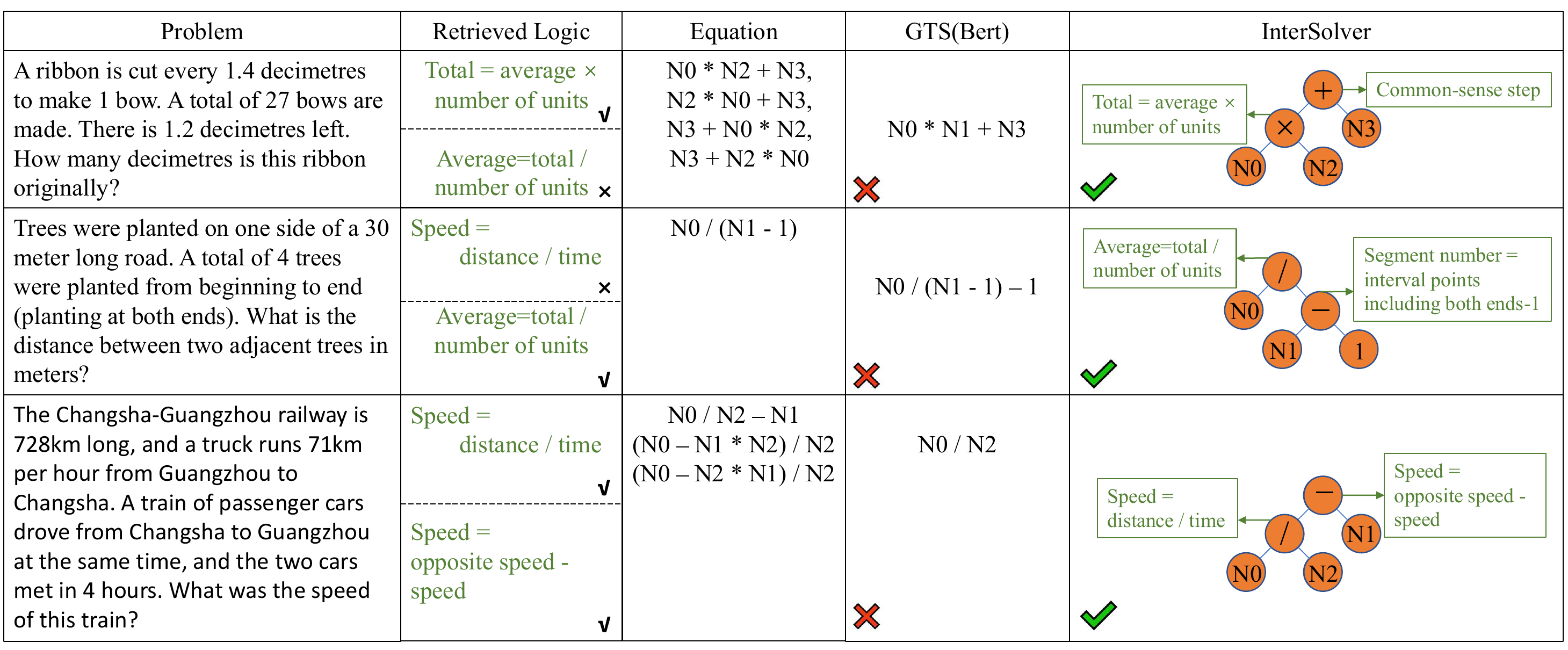}}
	\caption{Case study on GTS(BERT) and LogicSolver on InterMWP test set. Equation denotes the annotated complete solution set. (Note that the results are represented as infix traversal of expression trees which is more readable than prefix traversal.)}
	\vspace{-3mm}
	\label{fig:case}
\end{figure*}

\subsection{Performance of Interpretability}
We use the proposed Logic Generator to achieve interpretability and use logic accuracy to evaluate the performance. As shown in Table \ref{tab:mr}, our LogicSolver achieves 76.0\% on logic accuracy which is superior to all the other baselines. Notably, our LogicSolver can outperform GTS(BERT) which has the same backbone network by nearly 10\% benefiting from our logical prompt-enhanced learning, which helps the solver leverage logical knowledge better and makes inner-node (operator) representations more suitable for explanation generation. 

\subsection{Analysis on Different Expression Tree Size}
\begin{figure}[htbp] 
	\centerline{\includegraphics[width=0.9\linewidth]{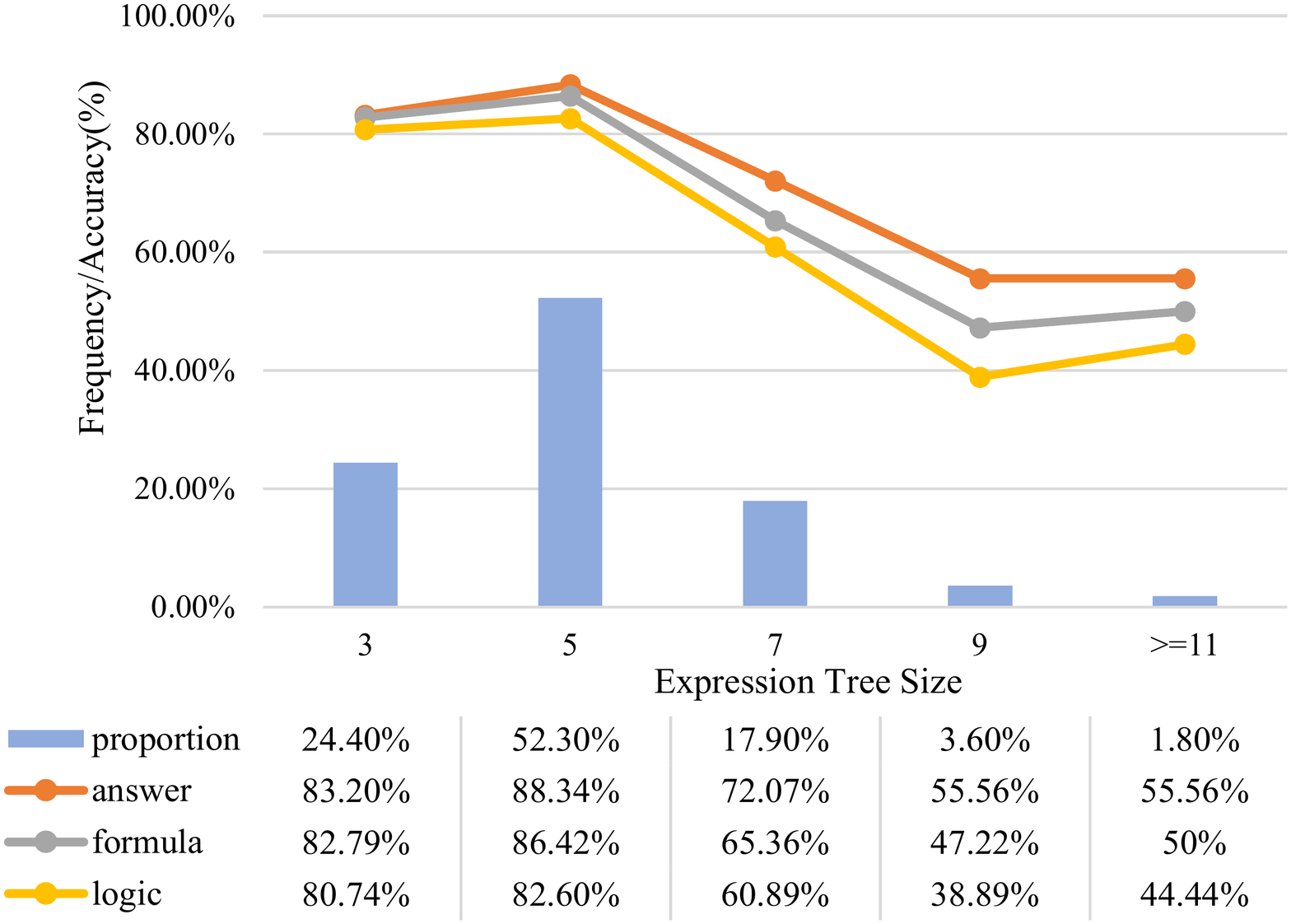}}
	\caption{Accuracy over different expression tree sizes. (\textit{proportion, answer, formula, and logic} denote data proportion, answer accuracy, formula accuracy, and logic accuracy over different expression tree sizes on the InterMWP test set.)}
	\vspace{-4mm}
	\label{fig:acc_length}
\end{figure}

We further evaluate the answer accuracy, formula accuracy, and logic accuracy on different problem expression tree sizes, as shown in Figure \ref{fig:acc_length}. We also show the corresponding data distribution. On the whole, the answer accuracy of problem solving decreases as the expression tree becomes longer, but the accuracy on tree size of 5 is higher than the tree size of 3 since the data proportion of tree size of 5 is obviously larger. In general, the longer the expression tree, the more difficult it is to be solved for both models and humans. 

\subsection{Case Study}
Finally, we conduct a case analysis and provide three cases in Figure~\ref{fig:case}. Benefiting from our logical prompt-enhanced learning on our InterMWP, our LogicSolver not only is more accurate in predicting operations, constants, and number words, but also can extract and generate correct logic reasoning procedures while GTS(BERT) is more likely to predict error expressions. In summary, our LogicSolver has gained a certain degree of interpretability while improving the accuracy of math word problem solving, showing the superiority of our InterMWP and LogicSolver. 

\section{Conclusion}
In this paper, to take a step towards interpretable MWP solving, we construct an interpretable math word problem dataset called InterMWP which consists of 11,495 MWP data and annotates interpretable logical formulas based on algebraic knowledge as the grounded linguistic logic of each solution equation. Different from existing MWP datasets, our InterMWP benchmark asks for a solver to not only output the solution expressions but also predict the corresponding logical formulas. We further propose LogicSolver  which is enhanced by logical prompts and is able to generate corresponding solution expressions and interpretable knowledge formulas in accord with the generated solution expressions, simultaneously. Experimental results show that our LogicSolver has stronger logical formula-based interpretability than baselines while achieving higher answer accuracy with the help of logical prompts, simultaneously.

\section{Limitations}
In this work, we make a step towards interpretable MWP solving by constructing a new MWP dataset called InterMWP and proposing a novel LogicSolver enhanced by logical prompts to infer out solution expressions and logical formula-based interpretability, simultaneously. However, there are still some limitations in our work. First, although our InterMWP is annotated with logical interpretability, the number of logical formulas are limited and needed to be extended for covering more cases when applying in real-world application. Second, even though our solver has better reasoning ability than current state-of-the-art methods on MWPs Solving and interpretation, it still needs more effort to design a more effective symbolic generation mechanism to enable a solver to handle more complex cases, such as more difficult problems with lager equations. 

\section*{Acknowledgements} 
This work was supported in part by National Key R\&D Program of China under Grant No. 2020AAA0109700, National Natural Science Foundation of China (NSFC) under Grant No.61976233 and Grant No.62206314, Guangdong Province Basic and Applied Basic Research (Regional Joint Fund-Key) Grant No.2019B1515120039, Guangdong Outstanding Youth Fund (Grant No. 2021B1515020061), GuangDong Basic and Applied Basic Research Foundation under Grant No.2022A1515011835, China Postdoctoral Science Foundation under Grant No.2021M703687, Shenzhen Fundamental Research Program (Project No. JCYJ20190807154211365), CAAI-Huawei MindSpore Open Fund, and The Open Project of Anhui Provincial Key Laboratory of Multimodal Cognitive Computation, Anhui University, No.MMC202107. We thank MindSpore for the partial support of this work, which is a new deep learning computing framwork\footnote{https://www.mindspore.cn/}.

\bibliography{acl2022}
\bibliographystyle{acl_natbib}


\newpage

\appendix
\label{sec:appendix}
\section{Dataset Statistics}
The InterMWP dataset consists of 11,495 problems and is divided into three parts randomly: 9495 training data, 1000 validation data, and 1000 test data. 
Table \ref{tab:lf} shows the statistics and some samples of four logic categories in InterMWP dataset. The contents in parentheses indicate the number of occurrences of the logical formulas of the category in the InterWMP dataset.

\begin{table}[htbp]
\centering
\resizebox{0.99\linewidth}{!}{
\begin{tabular}{l}
\hline
\textbf{Geometric Logics (988)} \\ \hline
$parallelogram\; area = bottom \times height$ \\ \hline
$rectangular\; area = length \times width $ \\ \hline
$square\; of\; the\; radius = radius \times radius$ \\ \hline
$circle\; area = PI\; \times square\; of\; the\; radius$ \\ \hline
$cuboid\; volume = bottom\; area \times height$ \\ \hline

\textbf{Physical Logics (4016)} \\ \hline
$speed = distance \div time$  \\ \hline
$distance = speed \times time$  \\ \hline
$time = distance \div speed$  \\ \hline
$workload = time \times work\; speed $ \\ \hline
$concentration = solute\; weight \div solution\; weight$  \\ \hline

\textbf{Financial Logics (1570)} \\ \hline
$expenses = price \times quantity$ \\ \hline
$insurance\; cost = insurance\; amount \times insurance\; rate$ \\ \hline
$sales\; income = cost + profit $ \\ \hline
$income\; after\; taxes = income\; before\; taxes - taxes$ \\ \hline
$taxes = tax\; payable \times tax\; rate$ \\ \hline

\textbf{Commonsense Logics (3852)} \\ \hline
$average = total \div number\; of\; units$  \\ \hline
$total = average \times number\; of\; units$ \\ \hline
$number\; of\; units = total \div average$ \\ \hline
$segment\; number = interval\; points\; excluding\; both\; ends + 1$ \\ \hline
$segment\; number = interval\; points\; including\; both\; ends - 1$ \\ \hline
\end{tabular}
}
\caption{Example logic formulas of different skills.}
\label{tab:lf}
\vspace{-3mm}
\end{table}

\begin{figure}[htbp]
	\vspace{-2mm}
	\subfigure[Problem length distribution]{
	  \begin{minipage}{0.99\linewidth}
		\centering
		\includegraphics[width=0.99\linewidth]{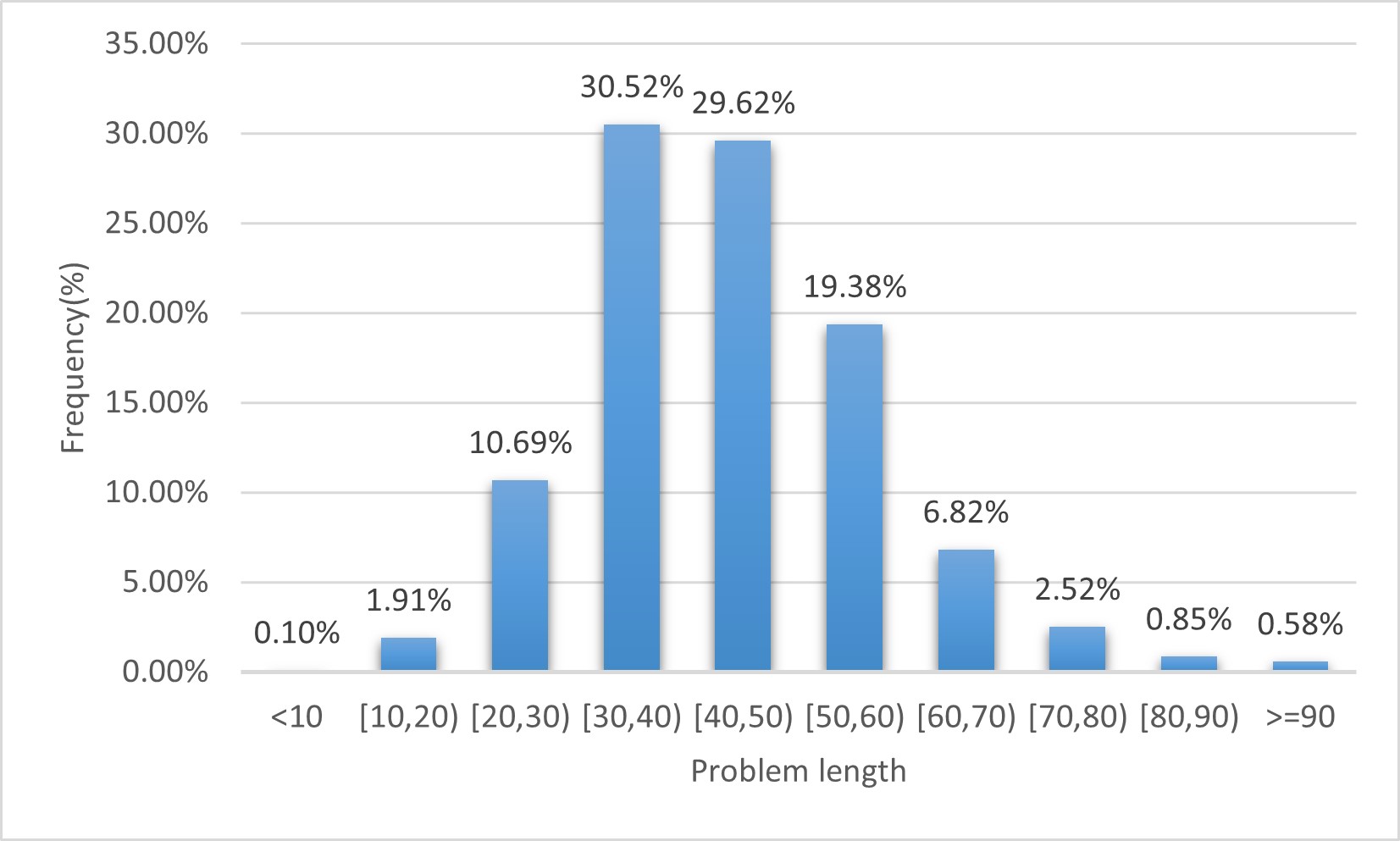}
		\label{fig:char_count}
		\vspace{-3mm}
	  \end{minipage}
	}
	\vspace{-2mm}
	\subfigure[Expression tree length distribution]{
	  \begin{minipage}{0.99\linewidth}
		\centering
		\includegraphics[width=0.99\linewidth]{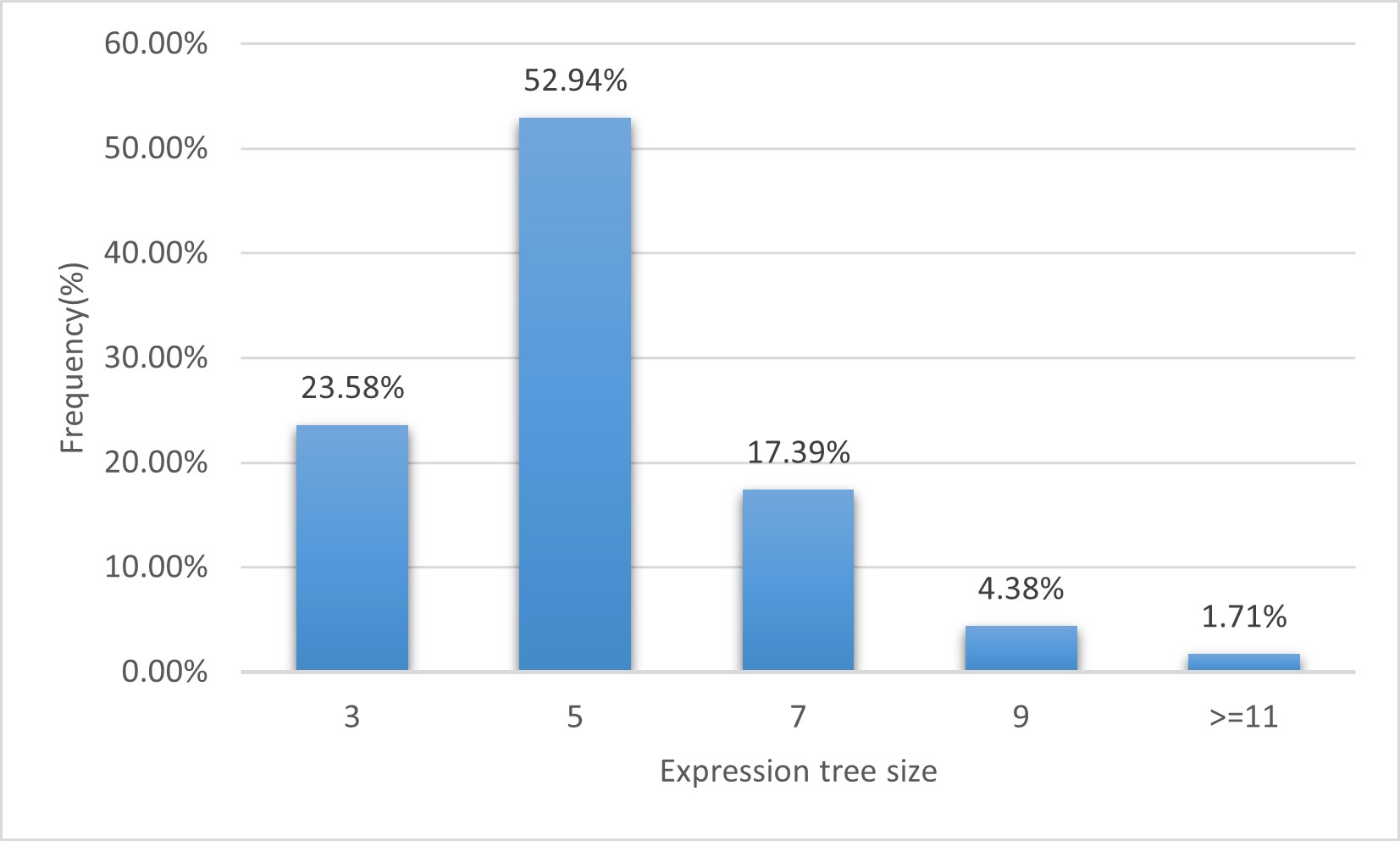}
		\label{fig:output_count}
		\vspace{-3mm}
	  \end{minipage}
	}
	\vspace{-2mm}
	\subfigure[Number of used logic formulas distribution]{
	  \begin{minipage}{0.99\linewidth}
		\centering
		\includegraphics[width=0.99\linewidth]{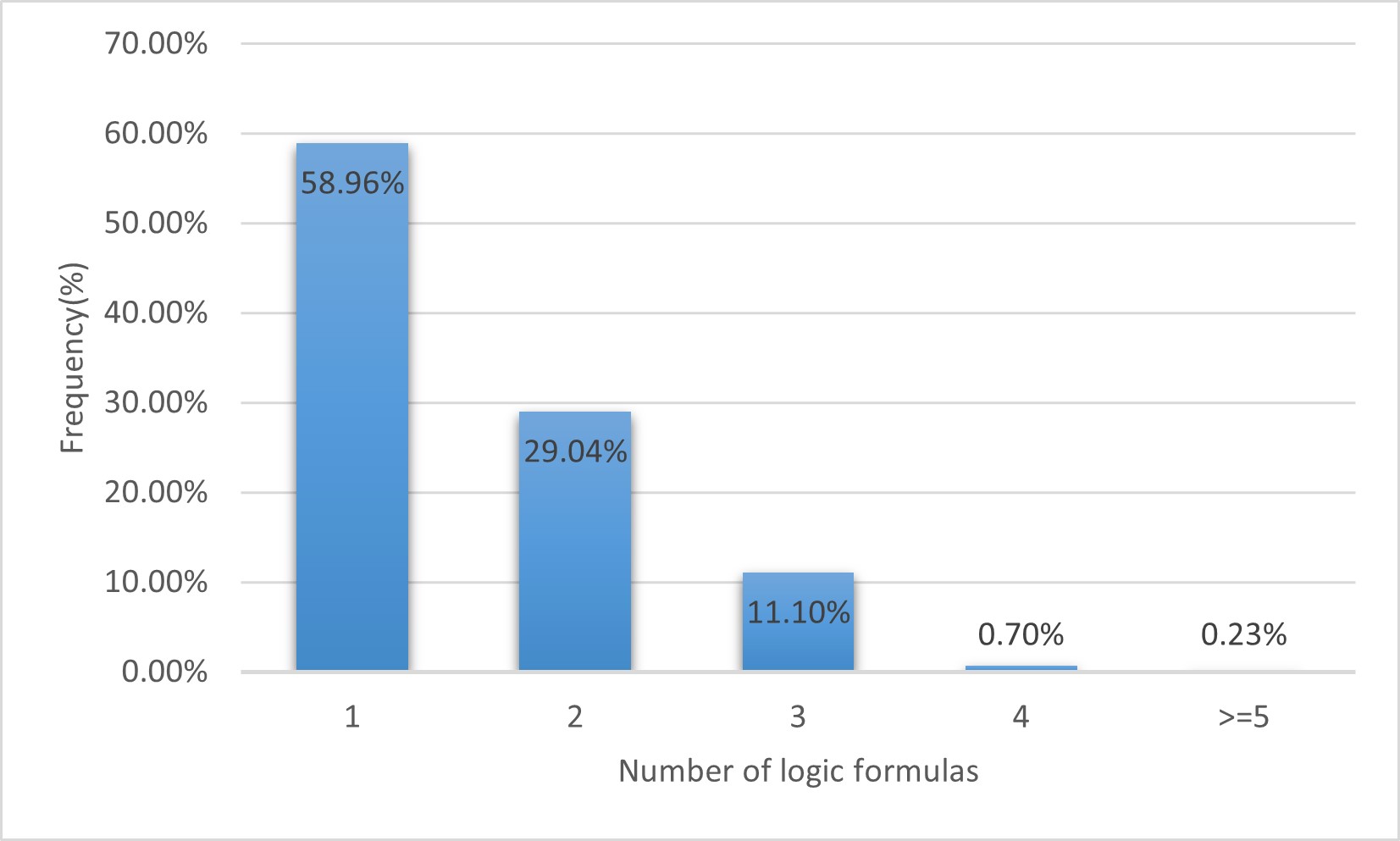}
		\label{fig:inter_count}
		\vspace{-3mm}
	  \end{minipage}
	}
	\caption{Dataset Statistics. We show the statistical characteristics of InterMWP (train+valid+test) for intuitive observation. We can observe that out InterMWP has moderate question length and expression size for MWP solving.}
	\label{fig:ds}
	\vspace{-3mm}
\end{figure}

The basic statistics of our InterMWP dataset are shown in Table \ref{tab:ds}. Figure~\ref{fig:ds} illustrates the distribution information about word-level question length, char-level question length, and expression tree length. For those problems with multi solutions, we take the shortest solution expression to count. From Figure~\ref{fig:ds}, we can observe that the lengths of most of the questions are adequate, which are not too long to understand for an MWP Solver. Besides, most expression tree contains less than 3 operators, which suggests that the questions should not very difficult to reason. However, the long tail in the distribution requires the MWP solvers to understand the complex mathematical relationships in the textual content.  

\begin{table}[htbp]
\centering
\small
\resizebox{0.99\linewidth}{!}{
\begin{tabular}{|c|c|c|c|c|}
\hline
 & Total & Train & Val & Test \\ \hline
Questions & 11,495 & 9,485 & 1,000 & 1,000 \\ \hline
Sentences &  16,308  & 13,456  & 1,408 & 1,444 \\ \hline
Words & 316,620 & 261,700 & 27,048 & 27,872 \\ \hline
\end{tabular}}
\caption{Basic statistics of our InterMWP dataset.}
\vspace{-3mm}
\label{tab:ds}
\end{table}

There are 210 algebraic knowledge formulas entailed in InterMWP. We list the most and least frequent knowledge formulas with a frequency greater than 5 in Table \ref{tab:fs}. It is shown that the distribution of formulas is not balanced but it is consistent with the real-world scene.

\begin{table}[htbp]
\centering
\resizebox{0.99\linewidth}{!}{
\begin{tabular}{|c|c|}
\hline
 formulas & \%  \\ \hline
Common-sense step & 56.35  \\ \hline  
average per unit = total number / number per unit & 4.75  \\ \hline 
total number = average number per unit $\times$ number of units & 4.75  \\ \hline 
number per unit = total number / average number per unit & 2.83  \\ \hline 
... & \\ \hline
increased price rate = 1 + price increment ratio  & 0.06  \\ \hline 
increased price = original price / increased price rate & 0.04  \\ \hline 
\end{tabular}}
\caption{Formulas statistics of our InterMWP dataset.}
\label{tab:fs}
\vspace{-4mm}
\end{table}

\section{Annotation Procedure}
\label{procedure}
Eighteen well-trained annotators with undergraduate degrees manually annotated solution equations with grounded algebraic knowledge formulas in the tree structure. Meanwhile, another annotator was required to summarize the algebraic knowledge formulas with the same meaning to eliminate logic redundancies. Finally, two annotators were asked to check the correctness of the annotated data from other annotators by conducting statistical sampling. 
When labeling the full solution to the test set, we use three operations to ensure the coverage of the full solution as far as possible: 1) The left and right sides of the symmetric operators (+,*) are recursively exchanged to generate new expressions; 2) Using sympy to obtain the simplified expressions and then operation 1 will be carried out; 3) New expressions are manually marked and then operation 1 and 2 are conducted.
If the correctness of an annotator's data is less than 96\% accurate, the data will be discarded.

\section{Implementation Details}
\label{details}
In our LogicSolver, the size of word embeddings and all hidden states for other layers are all set as 768, following the configuration of BERT-base~\cite{bert}. In each epoch, all training data is shuffled randomly and then cut into mini-batches. BERT models in LogicSolver are initialized by pre-trained BERT-wwm~\cite{chinese-bert-wwm} for Chinese. Our LogicSolver is optimized by ADAM optimizor~\cite{adam} with $\beta_1$ = 0.9, $\beta_2$ =0.999, and $\epsilon$ = $1e^{-8}$. The mini-batch size is set as 32, 32, and 32 for the retriever, encoder-decoder, and logic generator respectively. The initial fine-tuning learning rate is set as $1e^{-5}$ and $1e^{-4}$ for pre-trained BERT models and tree-decoder and then decreases to half every 25 epochs. To prevent overfitting, we set the dropout rate as 0.5 and weight decay as $1e^{-5}$. The training epochs are set as 20, 100, and 100 for retriever, encoder-decoder, and logic generator respectively. During solution expression generation, we use the beam search algorithm to generate expression trees and predict logic formulas.

\section{Experiments on logic formulas}
\label{exp_logic}
We also evaluate the formula accuracy and logic accuracy on the samples contain top-10 logic formulas in the test split of InterMWP dataset. The results are shown in Table \ref{tab:top_10_acc}. The performance gap of our LogicSolver relative to GTS(Bert) is significant on most logic formulas.

\begin{table}[htbp]
\small
\centering
\resizebox{0.99\linewidth}{!}{
\begin{tabular}{c|c|c}
\toprule
Logics & formula acc & logic acc \\
\midrule

\tabincell{c}{total number = average number per unit \\ $\times$ number of units} & 77.9/\textbf{78.6} & 53.6/\textbf{68.6} \\ 

\specialrule{0.01em}{3pt}{3pt}
\tabincell{c}{total number = number of units \\ $\times$  average number per unit} & 77.3/\textbf{78.7} & 53.9/\textbf{68.8} \\ 

\specialrule{0.01em}{3pt}{3pt}
\tabincell{c}{
average number per unit = total number \\ $\div$ number of units
} & 76.2/\textbf{78.6} & 54.8/\textbf{76.2}\\ 

\specialrule{0.01em}{3pt}{3pt}
expenses = quantity $\times$ price & 74.5/\textbf{76.5} & 47.1/\textbf{66.7} \\

\specialrule{0.01em}{3pt}{3pt}
\tabincell{c}{number of unit = total number \\ $\div$ average number per unit} & 76.8/\textbf{78.3} & 62.3/\textbf{71.0} \\ 

\specialrule{0.01em}{3pt}{3pt}
expenses = price $\times$ quantity & 74.0/\textbf{76.0} & 48.0/\textbf{68.0} \\

\specialrule{0.01em}{3pt}{3pt}
Working speed = workload $\div$ time& 71.4/\textbf{76.2} & 42.9/\textbf{71.4} \\ 

\specialrule{0.01em}{3pt}{3pt}
distance = speed $\times$ time & \textbf{74.1}/71.6 & \textbf{66.4}/65.2 \\ 

\specialrule{0.01em}{3pt}{3pt}
speed = distance $\div$ time & \textbf{76.7}/74.4 & 67.4/67.4 \\ 

\specialrule{0.01em}{3pt}{3pt}
Rectangle area = length $\times$ width & 56.4/56.4 & 41.0/\textbf{46.2} \\ 
\bottomrule
\end{tabular}
}
\caption{Formula accuracy and logic accuracy on the samples contain top-10 logic formulas with the most occurrences in the test split. (To the left of the semicolon `/' is the result of GTS(Bert), and to the right is the result of LogicSolver.)}
\vspace{-1mm}
\label{tab:top_10_acc}
\end{table}

\end{document}